\documentclass[conference, a4paper]{IEEEtran}

\usepackage[pagebackref=true,breaklinks=true,colorlinks,bookmarks=false]{hyperref}

\IEEEoverridecommandlockouts

	\usepackage[turkish]{babel}
	\usepackage[utf8]{inputenc}
	\usepackage[T1]{fontenc}
    \usepackage{amsfonts}

\usepackage{cite}

\ifCLASSINFOpdf
  \usepackage[pdftex]{graphicx}

\else

\fi

\usepackage[cmex10]{amsmath}

\usepackage{multirow}
\usepackage{array}
\usepackage{mathtools}
\DeclarePairedDelimiter\floor{\lfloor}{\rfloor}
\usepackage{amsmath}
\DeclareMathOperator*{\argmax}{arg\,max}

\usepackage[lofdepth,lotdepth]{subfig}

\AtBeginDocument{%
  \renewcommand\tablename{TABLE}
}

\AtBeginDocument{%
  \renewcommand\abstractname{Abstract}
}

\AtBeginDocument{%
  \renewcommand{\figurename}{Figure}
}

\begin{document}

\title{GRJointNET: Synergistic Completion and Part Segmentation on 3D Incomplete Point Clouds}

\author{Yiğit Gürses\\ \small Technical University of Munich 
        \and 
        Melisa Taşpınar\\ \small Bilkent University
        \and 
        Mahmut Yurt\\ \small Stanford University 
        \and 
        Sedat Özer\\ \small Özyeğin University}

\maketitle

\begin{abstract}
Segmentation of three-dimensional (3D) point clouds is an important task for autonomous systems. However, success of segmentation algorithms depends greatly on the quality of the underlying point clouds (resolution, completeness etc.). In particular, incomplete point clouds might reduce a downstream model's performance. GRNet is proposed as a novel and recent deep learning solution to complete point clouds, but it is not capable of part segmentation. On the other hand, our proposed solution, GRJointNet, is an architecture that can perform joint completion and segmentation on point clouds as a successor of GRNet. Features extracted for the two tasks are also utilized by each other to increase the overall performance. We evaluated our proposed network on the ShapeNet-Part dataset and compared its performance to GRNet. Our results demonstrate GRJointNet can outperform GRNet on point completion. It should also be noted that GRNet is not capable of segmentation while GRJointNet is. This study\footnote{This paper is the authors' enhanced version. This work was originally published at IEEE SIU 2021 and available on IEEE Xplore with DOI:10.1109/SIU53274.2021.9477918.}, therefore, holds a promise to enhance practicality and utility of point clouds in 3D vision for autonomous systems.

\end{abstract}
\begin{IEEEkeywords}
Point Clouds, Completion, Segmentation.
\end{IEEEkeywords}

\IEEEpeerreviewmaketitle

\IEEEpubidadjcol

\section{Introduction}
With the new developments in image acquisition technologies and the widespread use of 3D sensors, the demand for 3D object processing algorithms has also increased. Various algorithms have been developed recently for this purpose as in \cite{Lai}, \cite{Afham}, \cite{Yu}, \cite{Zhou}, as well as \cite{GRNet}, which forms the basis of our method. A particular relevant application is the processing of 3D point clouds which are often obtained with sensors such as Lidar \cite{Zhang}. 3D objects are commonly represented by 3D point clouds, as point clouds can effectively represent and describe the same scene with significantly lower data size \cite{pfnet} compared to voxel based methods. However, 3D point clouds obtained with sensors tend to be incomplete due to various factors such as light reflection, occlusion, low sensor resolution and limited viewing angles \cite{GRNet}, \cite{pfnet}, \cite{Snowflake}. Hence, the performance of algorithms that use the data as it is suffer \cite{Chen}. For this reason, a pre-processing step that implements some form of completion and resolution enhancement is often included \cite{pmp}. GRNet~\cite{GRNet} is one of the recently proposed deep learning-~based algorithms for this 3D point cloud completion. 

Part-segmentation is another type of vision task used in various domains, such as \cite{covidsegmentation}, \cite{plant} and \cite{car}, where each point is assigned one of the predefined labels to segment the whole object into smaller meaningful parts. However, segmenting an incomplete object where some parts may be wholly missing can be unproductive. In this context, completion algorithms can be used as an intermediary step before segmentation to obtain better results \cite{pmp}. However, such multi-step processes typically require more resources and can not be parallelised, leading to longer run-times. Therefore, a preferable alternative is to perform point cloud completion and segmentation jointly. In this study, we present a new architecture that aims to simultaneously complete and segment 3D incomplete point clouds, and we call this architecture GRJointNET. 

\begin{figure*}
  \centering
  \shorthandoff{=}
  \includegraphics[width=\textwidth]{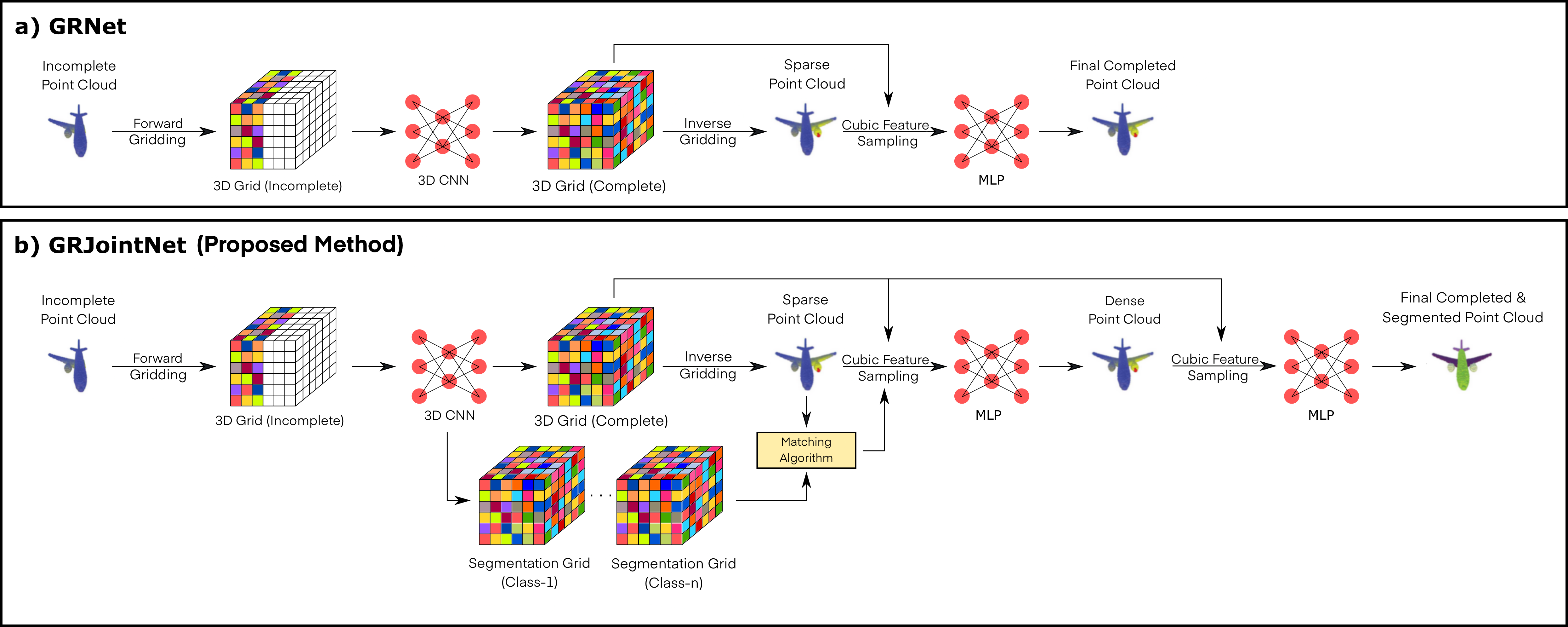}
  \caption{a) a) The architecture of the base method (GRNet \cite{GRNet}) and b) the architecture of our proposed method are shown. GRJointNet takes an incomplete point cloud as input and processes this cloud data through two different branches (completion and segmentation) to output a completed and segmented point cloud.} 
  \label{fig:sekil1}
\end{figure*}

GRJointNET makes use of 3D convolutional layers, three differentiable gridding layers (gridding, gridding reverse, and cubic feature sampling) from \cite{GRNet}, a novel segmentation reverse gridding layer and a novel synergistic feature sampling method (see Figure~\ref{fig:sekil1}). In this method, the incomplete regions in the input point clouds are completed and the points in the created point clouds are and segmented simultaneously.

Our main contributions can be summarized as follows:
\begin{itemize}
    \item While GRNet cannot perform segmentation together with completion, GRJointNet can perform both segmentation and completion synergistically.
    \item Unlike the GRNet architecture, our GRJointNet architecture uses segmentation estimates while performing incomplete point completion in the last layer.
    \item Comparative experimental results on the Shape-Net Part dataset are presented.
\end{itemize}

\section{Related Works}

Several recent studies have presented various deep neural network models to segment and complete 3D objects \cite{segcloud}, \cite{vox}, \cite{pointnet}, \cite{3DExtrapolation}. One of the studies that pioneered point cloud-based research in this field is PointNet \cite{pointnet}. Although this type of 3D point space-based models have demonstrated some success in the segmentation task, their performance depends on the completeness of the points in the point cloud~\cite{pfnet}. However, as mentioned previously, 3D point clouds tend to be incomplete for many reasons \cite{GRNet}, \cite{pfnet}, \cite{Snowflake}. In other words, when working on 3D point clouds, during applications such as segmentation, completing the incomplete point clouds first is considered a separate task.

Many recent and independent studies have successfully demonstrated that incomplete point clouds can be completed using deep neural architectures \cite{GRNet,pfnet,msn,atlasnet}. Some of these studies perform the completion process using multi-layer perceptrons (MLP) on raw point clouds \cite{pfnet}. However, such MLP-based methods have difficulty in exploiting spatial correlations between points due to the context-unaware architecture of MLPs. For this reason, newer studies have aimed to utilize 3D CNN's (convolutional neural networks) by voxelizing the point clouds. Even so, in such studies, performance decreases may be observed due to loss of geometric information during the voxelization process \cite{shape,high-res,voxsegnet}. A recent approach, GRNet \cite{GRNet}, proposes a model that represents point clouds with 3D grids with the aim of preserving geometric and structural information. Although GRNet is relatively successful at its purpose, it does not have segmentation capabilities.

In this study, we enhance the GRNet structure and present an end-to-end architecture that performs both completion and segmentation simultaneously. We call our architecture GRJointNET. GRJointNET, using GRNet as its base structure, is designed to improve the capabilities of GRNet by incorporating point cloud completion into its framework.

\begin{figure*}
    \centering
    \begin{minipage}{.48\textwidth}
        \centering
	    \shorthandoff{=} 
        \includegraphics[height=1.15\textwidth]{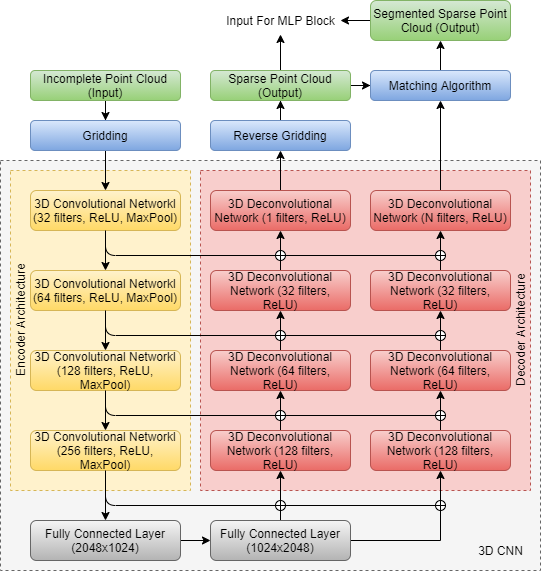}
    \end{minipage}%
    \begin{minipage}{.48\textwidth}
        \centering
        \shorthandoff{=} 
        \includegraphics[height=1.15\textwidth]{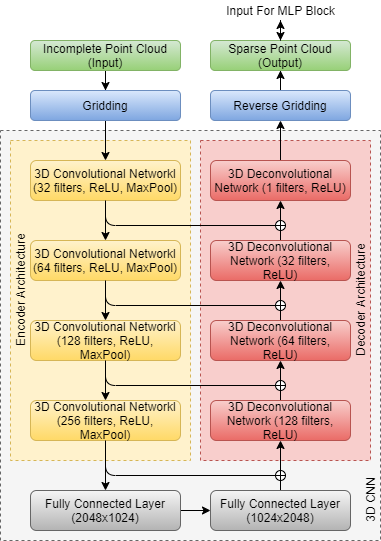}
    \end{minipage}
    \caption{\small Comparison of the 3D CNN structures of GRJointNet and GRNet. The figure on the left shows the 3D CNN structure used in GRJointNet, and the figure on the right shows the 3D CNN structure used in GRNet.\normalsize}
     \label{fig:GRNet}
\end{figure*}

\section{The Proposed Architecture}

The architecture of our proposed method (GRJointNet) is given in Figure~\ref{fig:sekil1}. In the GRJointNet architecture, there are five fundamental components including (i) gridding, (ii) gridding reverse, (iii) cubic feature sampling, (iv) the 3D convolutional neural network, (v) the multilayer perceptron, (vi) the mapping algorithm and (vii) the loss functions. 

Below, we explain each of those components.

\subsubsection{Gridding} 
It is not defined how to apply 2D and 3D convolutions directly on irregular point clouds, which is why placing the data on a 3D grid structure is a preferred method. Such methods are referred as voxelization. After voxelization, we can apply 2D and 3D convolution operations directly. However, since this process is not reversible, voxelization methods inherently lead to loss of geometric or semantic information. Therefore, in this study, we include a differentiable gridding layer to transform irregular 3D point clouds into regular 3D grids. The targeted 3D grid consists of $N^3$ individual vertices (where $N$ denotes the number of vertices on one dimension of the grid), covering the entire point cloud given as input and taking the shape of a regular cube. Each cell in this grid contains 8 different vertices, each with a weight value. The total number of vertices is $N^3$ with
\begin{equation}
    V=\{v_i\}_{i=1}^{N^3}, W=\{w_i\}_{i=1}^{N^3}, v_i=(x_i,y_i,z_i).
\end{equation}
Here, $W$ holds the cell values whereas the set $V$ holds the vertex coordinates of the corresponding cells. $v_i$ defines the 3D point at the $i^th$ index. If a point from the point cloud object lays within a cell with $8$ vertices, the weights of these vertices for that point is determined as follows:
\begin{equation}
    w_i^p=(1-|x_i^v-x|)(1-|y_i^v-y|)(1-|z_i^v-z|)
\end{equation}
Here, $x$ represents the projection of a sample coming from the point cloud onto the x-axis, $y$ represents its projection onto the y-axis and $z$ onto the z-axis. $x_i^v$, $y_i^v$ and $z_i^v$ define a vertex neighbouring the point in question. The final weight $w_i$ of the vertex is then calculated as follows: $ w_i = \sum_{p\in N(v_i)}\frac{w_i^p}{|N(v_i)|}$, 

where $N(v_i)$ is the set of points neighbouring the vertex $v_i$. The condition that a point $p$ neighbors $v_i$ can be written as $|x_i^v-x|<1,|y_i^v-y|<1,|z_i^v-z|<1$.

\subsubsection{Gridding Reverse}
Gridding reverse is the operation that creates the sparse point cloud from the given 3D grid. The points $p_i^s$ are calculated as follows:
\begin{equation}
    p_i^s=(\sum_{j\in N(v_i)}{w_jv_j})/({\sum_{j\in N(v_j)}{w_j}}),
\end{equation}
Here, $N(v_i)$ denotes the set of vertices neighboring $p_i^s$, $w_j$ denotes the weight of the $j$th vertex in $N(v_i)$, and $v_j$ denotes the spatial position of that vertex.

\subsubsection{Cubic Feature Sampling}
Classical MLP-based methods~\cite{pointnet} working on 3D point clouds suffer from global and local information loss between neighboring points because they do not take into account local spatial features. To solve this problem, we use the cubic feature sampling technique in our proposed method. This method collects relevant features from the grid for each point in the sparse point cloud. In short, the features of the eight neighboring vertices surrounding the point $p_i$ are combined and the input of the MLP ($o_p^i$) relative to that point is created as follows: $o_p^i = [p_i,f_1^i,f_2^i,...,f_8^i]$. Here, $o_p^i$ denotes the input of the MLP due to the point $p_i$, whereas $f_j^i$ denotes the feature map of the vertices surrounding $p_i$ from the 3D CNN. Note that the cubic feature sampling takes feature maps from the first three transposed convolutional layers in the 3D CNN, and it randomly samples 8 features from each channel per each point.

\subsubsection{3D Convolutional Neural Network}
Both GRNet and GRJointNet each contain a 3D CNN structure. The difference between these two 3D CNN structures can be seen comparatively on Figure~\ref{fig:GRNet}. The 3D CNN in the proposed approach contains an encoder-decoder structure. The encoder consists of four 3D convolutional layers, each of which includes a padding of 2, batch normalization, max pooling layers of kernel size 4, and a leaky ReLU activation. It is followed by fully connected layers of dimensions 1024 and 2048. Meanwhile, the decoder contains four transposed convolutional layers, each of which includes a padding of 2, stride of 1, a batch normalization, and a leaky ReLU activation. The general formulation of the 3D CNN is defined as follows: $W'=3DCNN(W)$; where $W$ is the output of the incomplete point cloud from the gridding process, and $W'$ is its completed version. Thus, the 3D CNN recovers the missing points in the given incomplete point cloud.

 \begin{figure*}[t!]
  \centering
  \shorthandoff{=}
  \includegraphics[width=0.9\textwidth]{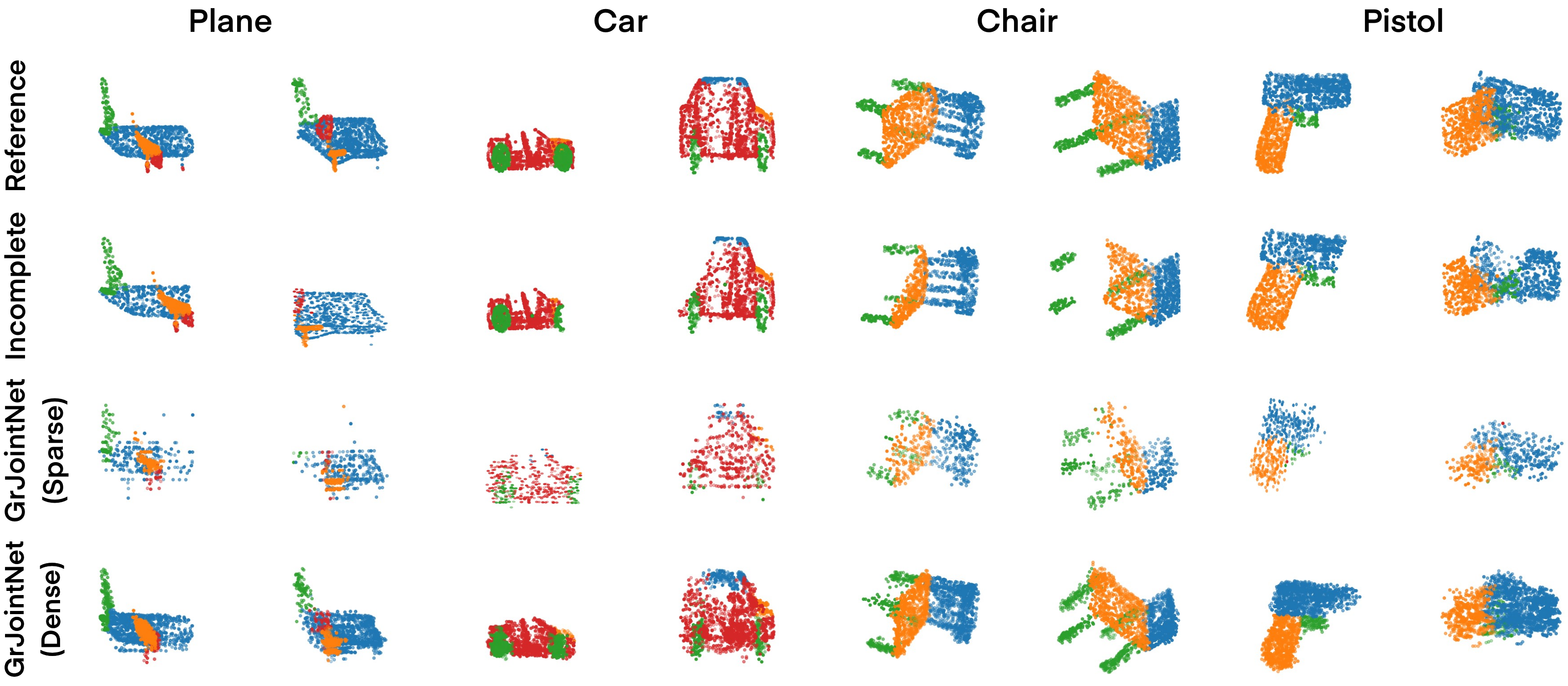}
  \caption{The results of incomplete point cloud completion performed by the proposed GRJointNet model are shown on incomplete, sparse and dense point clouds. Two samples were used for each of the following categories: plane, car, chair and pistol.}
  \label{gorselsekil}
\end{figure*}

\subsubsection{Multilayer Perceptron (MLP)}
The MLP architecture in the proposed method aims to recover fine details from the sparse point cloud by using the deviation between the final completed/segmented point cloud and the sparse point cloud. The MLP architecture encompasses four fully connected (FC) layers with sizes 12, 1000, 2000, and 3584, respectively.

\subsubsection{Mapping Algorithm}
The performance of GRJointNet depends on the efficient use of the deconvolutional layers that form the segmentation grid to learn well. For this purpose, we segmented the sparse point cloud and used this segmentation in back-propagation with cross entropy loss. The mapping algorithm works as follows:
\begin{equation}
\begin{array}{cc}
     c^p_x= \floor{N(p_x+1)}, c^p_y= \floor{N(p_y+1)},c^p_z= \floor{N(p_z+1)}, \\
     \text{and }   b^p= \argmax_n BI_n[c^p_x,c^p_y,c^p_z].
\end{array}
\end{equation}

Here $c^p_x, c^p_y$ and $c^p_z$ indicate the indices of the cell that point $p$ will fall into in a segmentation grid of size $N^3$. $BI_n$ denotes the $n^{th}$ of the resulting $n$ segmentation grids and contains the spatial probabilities of the segmentation category numbered $n$. $b^p$ indicates the segmentation category assigned to point $p$ at the end of the mapping algorithm.

\subsubsection{Loss Functions} 
The {\it Chamfer distance} between the actual ground truth and the completed/segmented objects is defined as:
\begin{equation}
    \small L_{CD} = \frac{1}{n_G}\sum_{g\in G}^{}\min_{m\in M} ||g-m||_2^2+\frac{1}{n_M}\sum_{m\in M}^{}\min_{g\in G} ||g-m||_2^2
\end{equation}
\normalsize
For each point in $G$, the closest point in $M$ is calculated based on the distance $L_2$. This $L_2$ distance is included in the loss. The same process is repeated for each point in $M$. For the segmentation loss, {\it cross entropy loss} was used.
\begin{equation}
    \small L_{CE} = -\sum^{C}_{i}t_{i}log\left(\frac{e^{s_{i}}}{\sum^{C}_{j}e^{s_{j}}}\right)
\end{equation}
\normalsize

Given that complete point clouds do not have ground truths for segmentation when they are first created; the ground truths are calculated using the original complete point cloud which has segmentation labels available. Each point in the generated clouds is assigned to the segmentation label of the point closest to it in the complete cloud. Afterwards, the segmentation predictions on both sparse and dense point clouds are compared to the ground truths we generated using cross entropy. Using only the Chamfer distance as a loss function to train GRNet is insufficient to check whether the predicted points match the geometry of the object. For this reason, networks that use only Chamfer distance tend to give an average shape that minimizes the distance of input and output points. This in turn causes a loss of information regarding the details of the object in question. Since point clouds are unsorted, it becomes difficult to apply $L_1$ / $L_2$ loss function or cross entropy directly on them. However, the gridding method introduced by GRNet \cite{GRNet} overcomes this problem by converting unsorted 3D point clouds into 3D grids. Therefore, GRNet introduces a novel loss function called {\it Grid Loss Function}. This loss function is defined as the distance $L_1$ between two sets of values of 3D grids. In other words:
\begin{equation}
    \small L_{Gridding}(W^{pred},W^{gt}) = \frac{1}{ {N^3}_{G}} \sum || W^{pred} - W^{gt}||.
\end{equation}
\normalsize
Here, $W^{pred},W^{gt}\in \mathbb{R}^{N_G^3}$. $ \small G_{pred}=<V^{pred},W^{pred}>$ and $\small G_{gt} = <V^{gt},W^{gt}>$ are 3D grids obtained by applying gridding to the ground truth ($G_{gt}$) and the predicted ($G_{pred) }$) point clouds. Additionally, N\textsubscript{G} corresponds to the resolution of the 3D grids. The last used loss function ($ L$) on the other hand is defined as follows: $ L = L_{CD} + L_{CE} + L_{Gridding}$.

\section{Experiments}
The performances of GRNet and GRJointNet were compared for four selected categories on the ShapeNet-Part dataset ~\cite{shapenetpart}. Given that GRNet is an algorithm designed to perform completion, we carried out our experiments separately for both completion and segmentation purposes. All algorithms were trained over 50 epochs. Adam optimization was used on both networks. In the completion experiments, a total of 11705 training samples and 2768 test samples were used from the ShapeNet-Part dataset. The results are shown comparatively in Table~\ref{tablo1} over four randomly selected individual classes including "car", "plane", "chair" and "pistol". In the table, we used the average Chamfer distance as the metric for performance comparison, where the smaller values are the better results and the best results are shown in bold. For each value pair $cd_{sparse} / cd_{dense}$ in Table~\ref{tablo1}, $cd_{sparse}$ and $cd_{dense}$ refer to the Chamfer distances of the sparse and dense completed point clouds to the ground truth, respectively.
 
 \renewcommand{\arraystretch}{1}
\begin{table}[t!]
\centering
\caption{\textsc{GRNet vs GRJointNet Results}}
\resizebox{0.9\columnwidth}{!}{%
\begin{tabular}{|c|c|c|}
\hline

                          & GRNet                & GRJointNet               \\ \hline
car                       &  6.26 / \textbf{2.92}    &     \textbf{6.18} / 3.00                  \\ \hline
plane                  &  5.70 / \textbf{1.49}     &      \textbf{3.27} / 1.50                 \\ \hline
chair                     &  5.52 / 2.92     &      \textbf{4.58} / \textbf{2.36}                \\ \hline

pistol                   &  13.07 / \textbf{1.83}    &      \textbf{12.71} / 1.85                \\ \hline
\end{tabular}
}
\label{tablo1}
\end{table}
 
In the part-segmentation experiment, since GRNet does not have a segmentation feature, we present only our results in the Figure~\ref{gorselsekil} using two examples from four different categories. In the figure, the first row shows the reference images, the second row shows the inputted incomplete point clouds, whereas the third and fourth rows respectively show the sparse and dense point clouds that are the outputs of the model, all together with the segmentation results.

\section{Conclusion}
In this study, a synergistic deep learning-based method is proposed for the completion and segmentation of incomplete (3D) point clouds. While the proposed method achieves near or better performance than our baseline method (GRNet) in the completion category, it can also successfully segment the completed point cloud to provide further functionality. In real-world autonomous system applications, the collected data is often incomplete and noisy while including data from several different types of sensors. In this context, it can be said that models focusing only on one task will be less efficient and perform worse compared to integrated systems that process all the available data synergistically. To that end, similar to the method proposed in this study, more useful and effective models that can use various features of the gathered data (position, distance, image, etc.) to perform multiple autonomous system-based tasks are being developed \cite{RLTracking, DetectionPaper,ozer2022siamesefuse}. Integrated systems using such mentioned methods allow use of common inputs at different components and as such, they can lower the required computational resources, while acquiring extra information from other components' internal processes to enhance each others performances.

\section*{Acknowledgment}
{\small This paper has been produced benefiting from the 2232 International Fellowship for Outstanding Researchers Program of TÜBİTAK (Project No:118C356). However, the entire responsibility of the paper belongs to the owner of the paper. The financial support received from TÜBİTAK does not mean that the content of the publication is approved in a scientific sense by TÜBİTAK.}

\renewcommand{\refname}{References} 
\bibliographystyle{ieeetr}
\bibliography{References} 

\end{document}